\documentclass[letterpaper]{article} 
\usepackage{aaai2026}  
\usepackage{times}  
\usepackage{helvet}  
\usepackage{courier}  
\usepackage[hyphens]{url}  
\usepackage{graphicx} 
\usepackage{natbib}  
\usepackage{caption} 
\usepackage{algorithm}
\usepackage{algorithmic}
\usepackage{amsmath}
\usepackage{booktabs}
\usepackage{subcaption}

\usepackage{newfloat}
\usepackage{listings}
\DeclareCaptionStyle{ruled}{labelfont=normalfont,labelsep=colon,strut=off} 
\floatstyle{ruled}
\newfloat{listing}{tb}{lst}{}
\floatname{listing}{Listing}
\title{Activations as Features: Probing LLMs for Generalizable Essay Scoring Representations}

\author {
    Jinwei Chi\textsuperscript{\rm 1},
    Ke Wang\textsuperscript{\rm 2}\thanks{Corresponding author.},
    Yu Chen\textsuperscript{\rm 2},
    Xuanye Lin\textsuperscript{\rm 3},
    Qiang Xu\textsuperscript{\rm 2}
}
\affiliations {
    \textsuperscript{\rm 1}Guangdong Institute of Smart Education, Jinan University, Guangzhou, China\\
    \textsuperscript{\rm 2}College of Information Science and Technology, Jinan University, Guangzhou, China\\
    \textsuperscript{\rm 3}Future Tech College, South China University of Technology, Guangzhou, China\\
    \{202434261005cjw, biabuluo, 1002xu\}@stu2024.jnu.edu.cn, 
    wangke@jnu.edu.cn, 
    202264700324@mail.scut.edu.cn
}

\begin{document}

\maketitle

\begin{abstract}
Automated essay scoring (AES) is a challenging task in cross-prompt settings due to the diversity of scoring criteria. While previous studies have focused on the output of large language models (LLMs) to improve scoring accuracy, we believe activations from intermediate layers may also provide valuable information. To explore this possibility, we evaluated the discriminative power of LLMs’ activations in cross-prompt essay scoring task. Specifically, we used activations to fit probes and further analyzed the effects of different models and input content of LLMs on this discriminative power. By computing the directions of essays across various trait dimensions under different prompts, we analyzed the variation in evaluation perspectives of large language models concerning essay types and traits. Results show that the activations possess strong discriminative power in evaluating essay quality and that LLMs can adapt their evaluation perspectives to different traits and essay types, effectively handling the diversity of scoring criteria in cross-prompt settings.
\end{abstract}


\section{Introduction}
Automatic essay scoring (AES) aims to score the characteristics of the article according to the given prompts. A reliable automatic essay scoring system can not only effectively reduce the burden of manual scoring, but also significantly improve the scoring efficiency, so this field has attracted much research attention.

Due to the difficulty of obtaining reliable ground-truth labels for essay data, recent research in AES has shifted toward cross-prompt scoring, expecting to develop models that consistently evaluate essays written in response to diverse prompts. However, essay scores are highly sensitive to writing prompts, and models often fail to generalize well across different prompts. Some studies have approached cross-prompt scoring as a domain adaptation problem, treating each prompt as a separate domain and applying transfer learning techniques to adapt models from source to target prompts \cite{phandi2015flexible, cummins2016constrained}. However, these methods need to use the data of the target domain, so the application is limited in cases of limited resources or new prompts. To solve this limitation, some researchers have tried to adopt methods such as comparative learning, meta learning, and knowledge alignment strategy based on clustering \cite{jiang2023improving, chen2023pmaes, chen2024plaes, li2025kaes}, to learn general features that do not depend on specific prompts. 

Many generative AI models have emerged with the wide application of the transformer architecture, showing strong language understanding ability. Some researchers began to explore the possibility of using large language models for automated essay scoring. The most common approaches are to directly input the natural language instructions for the tasks we expect LLMs to perform \cite{malik2024exploring, helmeczi2023few}, or to fine-tune open-source LLMs \cite{mazzullo2025fine}. However, these approaches only use the final output of the LLMs and do not fully leverage the activations of LLMs that contain rich semantic information.

We believe that semantically rich activations can provide useful information in cross-prompt automated essay scoring tasks. To validate this idea, we used linear and nonlinear probes to fit the activations. Probes are a commonly used analytical tool that trains a simple downstream classifier based on a frozen model representation to test whether the activation contains information that is discriminative for the target task. Results show that even simple linear probes achieved strong performance in cross-prompt settings, indicating that LLMs already encode discriminative essay quality information in their activations. Furthermore, the significant performance gap between linear and nonlinear probes further suggests that this information is largely represented linearly. To gain a deeper understanding of the factors that influence the discriminative power of LLM activations, we conducted a systematic comparison across six language models and four types of input content (hereafter referred to as input content to differentiate between prompts given to the LLMs and the essay prompts), analyzing how variations in model and input design affect the encoding of trait-relevant information. Additionally, by computing the directional representations of essays under different prompts across various trait dimensions, we found that LLMs adapt their evaluation perspectives based on essay type and trait, demonstrating flexibility in handling diverse scoring criteria.

Our contributions are as follows:
\begin{itemize}
    \item We propose to explore the internal representations of large language models for cross-prompt automated essay scoring, shifting the focus from output predictions to intermediate-layer activations.
    \item We fit linear probes using activations of LLMs, and find that the linear probe achieves performance surpassing existing methods on certain attention heads, demonstrating the potential of internal LLM representations in capturing essay quality.
    \item We analyze the directions corresponding to different traits and prompts, and find that these directions differ, revealing that LLMs can adapt their evaluation perspectives to various traits and essay types.
\end{itemize}

\section{Related Work}

\subsection{Automated Essay Scoring}
Automated essay scoring (AES) is a long-standing and active research topic, with early research focusing on holistic scoring tasks. Traditional approaches rely heavily on classical machine learning techniques where domain experts manually extract linguistic, syntactic, and semantic features from essays to train regression or classification models. Some work \cite{amorim2018automated, ridley2020prompt, uto2020neural} tried to design and acquire some handcrafted features for automated essay scoring. Since the handcrafted features are independent of prompts, these features are also widely used in cross-prompt essay scoring. However, the obvious problem is that most handcrafted features focus on the shallow linguistic level, such as word and sentence level statistics, which cannot effectively represent the higher linguistic features, such as semantic coherence, logical structure, and quality of argumentation in the essay.

The rise of deep learning has shifted research from holistic scoring to multi-feature and cross-prompt tasks. Neural models can automatically extract high-level features from raw text, reducing the need for manual feature engineering and improving robustness across prompts. \cite{xie2022automated} proposed a two-input model to learn correlations between known and unknown essays. \cite{do2023prompt} introduced prompt and trait attention, integrating handcrafted features and LDA-based coherence to generalize to unseen prompts.

Recent advances in LLMs have opened new directions for AES. These models, pre-trained on massive corpora, demonstrate strong language understanding and generation capabilities. \cite{cai2025rank} fine-tuned LLMs using trait-enhanced paired data to judge relative rankings between target and reference essays based on key scoring dimensions. \cite{boquio2024beyond} leveraged Bert's middle-layer semantic knowledge combined with pooling strategies to enhance model performance. Building on these approaches, we explore using LLM activations containing rich semantic information for cross-prompt automated essay scoring.

\subsection{Probing}
Probing refers to a supervised method for uncovering the feature representations of specific concepts learned by a trained neural network within the activation space of that network. A probe is a simple, linear model trained to predict ground-truth labels. It is also known as a linear probe. The reason for using linear probes is that it is hoped that concepts follow a linear hypothesis \cite{mikolov2013efficient} in the representation space, that is, concepts are linearly represented as directions in the representation space.

Most of the existing research work \cite{wang2024inferaligner, wang2025adaptive} on the internal concept representations of LLMs advocates detecting the outputs of individual attention heads and training a probe for each attention head. Take Llama-2-7b-chat-hf as an example. Llama-2-7b-chat-hf has 32 layers, and each layer has 32 heads. Therefore, 32×32=1024 probes need to be trained.

Existing studies have demonstrated, using linear probing methods, that certain high-level concepts, such as positive and negative sentiment \cite{hollinsworth2024language}, space and time in the real world \cite{gurnee2023language}, language \cite{bricken2023towards}, and safety \cite{arditi2024refusal}, can be linearly represented within large language models. These findings suggest that large language models automatically learns how to encode abstract semantics and complex cognitive structures during training, and that there is some near-linear dimension in their internal embedding space that can correspond to human interpretable concepts. Previous research has focused on abstract or generic concepts, and whether writing quality is similarly linearly encoded in LLMs remains open. We therefore attempt to use probing techniques on cross-prompt automated essay scoring task.

\section{Methodology}

\subsection{Input Content}
To allow LLMs to evaluate essays and obtain activations that are only relevant to the trait of the essay, we designed a simple input content template for LLMs. The template consists of three parts: prompt, essay, and instruction. The instruction guides LLMs in evaluating the essay from a specified perspective.

\begin{figure}[h]
    \centering
    \includegraphics[width=\linewidth, trim=0 210 0 210, clip]{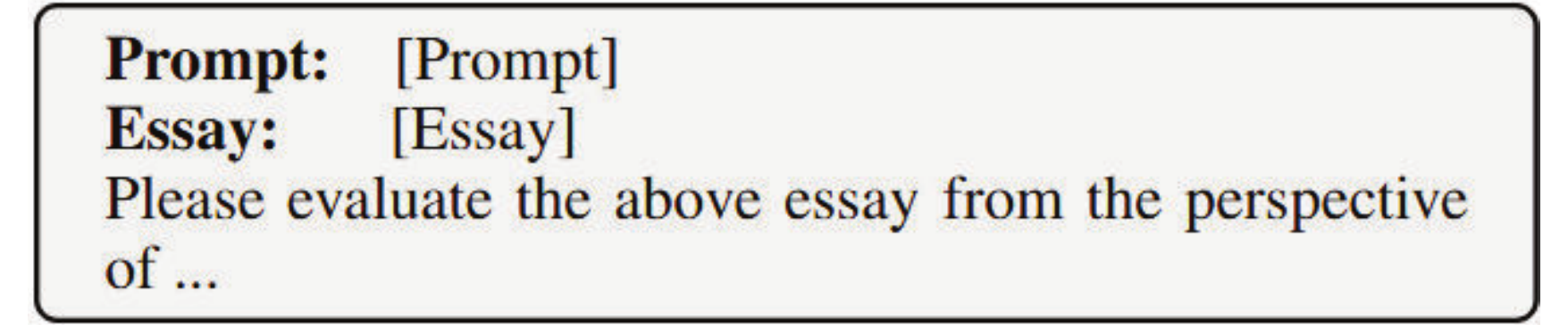}
    \caption{Input content template.}
    \label{template}
\end{figure}

For each trait, including the holistic score, we designed different instructions. All of which are shown in the \textit{Appendix A}.

\subsection{Acquisition of Activations}
We extracted the output vector of each attention head at the final token position of the input sequence. This position is widely regarded as capturing the model’s semantic understanding of the entire input, thus providing a highly abstract representation.

\subsection{Probe}
After obtaining the activations, we trained probes to predict the scores of the essays. We used the standard probing technique \cite{alain2018understanding, belinkov2022probing, yang2025qwen3}, which fits a simple model on the network activations to predict some target label associated with labeled input data. We fitted linear ridge regression probes:

\begin{equation}
\begin{aligned}
\boldsymbol{y}_{l,h} &= \boldsymbol{X}_{l,h} \hat{\boldsymbol{W}}, \\
\text{where} \quad \hat{\boldsymbol{W}} &= \left( \boldsymbol{X}_{l,h}^\top \boldsymbol{X}_{l,h} + \lambda \boldsymbol{I} \right)^{-1} \boldsymbol{X}_{l,h}^\top \boldsymbol{y}
\end{aligned}
\tag{1}
\end{equation}

where $\lambda$ = 0.01, is the regularization parameter, $\hat{\boldsymbol{W}}$ is the weight vector, $\boldsymbol{X}_{l,h}$ is the activations of head $\boldsymbol{h}$ of layer $\boldsymbol{l}$ and $\boldsymbol{y}_{l,h}$ is the final prediction of head $\boldsymbol{h}$ of layer $\boldsymbol{l}$. Since the labels were normalized to $[0, 1]$, we clipped the probe outputs accordingly: values below 0 were set to 0, values above 1 to 1.

\section{Experiment}

\subsection{Dataset}
We experimented with the same dataset as the baseline system, which comprises the publicly available Automated Student Assessment Prize (ASAP) and ASAP++ datasets. The ASAP dataset contains eight different essay sets, with essays in each set responding to a different prompt. Each essay is awarded a human-rated score for the holistic quality of the essay, and the essays for prompts 7 and 8 are additionally assigned scores for some relevant traits according to a scoring rubric.

Since only prompts 7 and 8 have trait scores, we also used the ASAP++ dataset, which builds on the original ASAP dataset, to provide scores for the various traits associated with prompts 1-6 to complement the original ASAP score. It should be noted that some traits only appeared in one prompt (e.g., \textit{Style} in prompt 7 and \textit{Voice} in prompt 8). Following \cite{ridley2021automated}, we excluded these traits. The dataset summary can be found in \textit{Appendix B}.

\subsection{Validation and Evaluation}
Based on previous work on cross-prompt automated essay scoring, we adopted a prompt-wise cross-validation approach \cite{jin2018tdnn, ridley2020prompt, ridley2021automated}. Specifically, essays from one prompt were used as the test set, while essays from the remaining prompts were used for training. For the evaluation, we used Quadratic Weighted Kappa (QWK), the official metric for ASAP competition and most frequently used for AES tasks, which measures the agreement between the human rater and the system. At training time, the training labels are normalized, meaning that the training label values are all between 0 and 1. At test time, the test labels use the original data. We trained a separate probe on the activations of each attention head in the model and reported the results of the head that achieved the best performance on the test set.

\subsection{Baselines}
For cross-prompt automated essay scoring, we employ six baseline systems: PAES \cite{ridley2020prompt}, CTS \cite{ridley2021automated}, ProTACT \cite{do2023prompt}, PMAES \cite{chen2023pmaes}, PLAES \cite{chen2024plaes} and KAES \cite{li2025kaes}. We used different LLMs, such as Llama-2-7b-chat-hf \cite{touvron2023llama}, DeepSeek-R1-Distill-Llama-8B \cite{guo2025deepseek}, Qwen3 Series \cite{yang2025qwen3}:Qwen3-8B, Qwen3-4B, Qwen3-1.7B, Qwen3-0.6B, to get results using our method.

\begin{table*}[t]
\centering
\setlength{\tabcolsep}{5.2pt} 
\begin{tabular}{lccccccccc}
\toprule
\textbf{Model} & \textbf{P1} & \textbf{P2} & \textbf{P3} & \textbf{P4} & \textbf{P5} & \textbf{P6} & \textbf{P7} & \textbf{P8} & \textbf{AVG} \\
\midrule
PAES \cite{ridley2020prompt} & 0.615 & 0.582 & 0.570 & 0.606 & 0.634 & 0.545 & 0.356 & 0.447 & 0.543 \\
CTS \cite{ridley2021automated} & 0.620 & 0.555 & 0.631 & 0.620 & 0.613 & 0.548 & 0.384 & 0.504 & 0.560 \\
PMAES \cite{chen2023pmaes} & 0.656 & 0.555 & 0.615 & 0.623 & 0.634 & 0.593 & 0.406 & 0.533 & 0.566 \\
PLAES \cite{chen2024plaes} & 0.648 & 0.563 & 0.604 & 0.623 & 0.634 & 0.593 & 0.403 & 0.533 & 0.575 \\
ProTACT \cite{do2023prompt} & 0.647 & 0.587 & 0.623 & 0.632 & 0.674 & 0.574 & 0.446 & 0.541 & 0.592 \\
KAES \cite{li2025kaes} & 0.623 & 0.611 & 0.619 & 0.636 & 0.685 & 0.593 & 0.433 & 0.577 & 0.597 \\
\midrule
\textbf{Qwen3-0.6B} & 0.610 & 0.619 & 0.598 & 0.662 & 0.681 & 0.621 & \underline{\textbf{0.570}} & 0.610 & 0.622 \\
\textbf{Qwen3-1.7B} & 0.649 & 0.650 & 0.629 & 0.672 & 0.693 & 0.645 & 0.517 & 0.564 & 0.627 \\
\textbf{Qwen3-4B} & 0.645 & 0.663 & 0.651 & 0.696 & \underline{\textbf{0.707}} & 0.658 & 0.456 & 0.565 & 0.630 \\
\textbf{Qwen3-8B} & 0.664 & \underline{\textbf{0.679}} & \underline{\textbf{0.676}} & \underline{\textbf{0.718}} & 0.701 & \underline{\textbf{0.678}} & 0.521 & 0.598 & \underline{\textbf{0.654}} \\
\textbf{DeepSeek-R1-Distill-Llama-8B} & 0.648 & 0.642 & 0.651 & 0.694 & 0.700 & 0.657 & 0.478 & 0.578 & 0.631 \\
\textbf{Llama-2-7b-chat-hf} & \underline{\textbf{0.665}} & 0.663 & 0.667 & 0.695 & 0.701 & 0.656 & 0.558 & \underline{\textbf{0.614}} & 0.652 \\
\bottomrule
\end{tabular}
\caption{\label{tab1}Main results on each prompt. The highest score in each column is bolded and underlined.} 

\smallskip
\smallskip
\smallskip

\centering
\setlength{\tabcolsep}{3pt}  
\begin{tabular}{@{}lcccccccccc@{}} 
\toprule
\textbf{Model} & \textbf{Holistic} & \textbf{Content} & \textbf{Org} & \textbf{WC} & \textbf{SF} & \textbf{Conv} & \textbf{PA} & \textbf{Lang} & \textbf{Nar} & \textbf{AVG} \\
\midrule
PAES \cite{ridley2020prompt} & 0.657 & 0.539 & 0.414 & 0.531 & 0.536 & 0.357 & 0.570 & 0.531 & 0.605 & 0.527 \\
CTS \cite{ridley2021automated} & 0.670 & 0.555 & 0.458 & 0.557 & 0.545 & 0.412 & 0.565 & 0.536 & 0.608 & 0.545 \\
PMAES \cite{chen2023pmaes} & 0.671 & 0.567 & 0.481 & 0.584 & 0.582 & 0.421 & 0.584 & 0.545 & 0.614 & 0.561 \\
PLAES \cite{chen2024plaes} & 0.673 & 0.574 & 0.491 & 0.579 & 0.580 & 0.447 & 0.601 & 0.554 & 0.631 & 0.570 \\
ProTACT \cite{do2023prompt} & 0.674 & 0.596 & 0.518 & 0.599 & 0.585 & 0.450 & 0.619 & 0.596 & 0.639 & 0.586 \\
KAES \cite{li2025kaes} & 0.679 & 0.594 & 0.543 & 0.636 & 0.574 & 0.476 & 0.611 & 0.595 & 0.641 & 0.594 \\
\midrule
\textbf{Qwen3-0.6B} & 0.689 & 0.620 & 0.565 & 0.610 & 0.605 & \underline{\textbf{0.549}} & 0.637 & 0.632 & 0.634 & 0.615 \\
\textbf{Qwen3-1.7B} & 0.688 & 0.619 & 0.559 & 0.639 & 0.654 & 0.529 & 0.640 & 0.635 & 0.664 & 0.625 \\
\textbf{Qwen3-4B} & 0.690 & 0.636 & 0.529 & 0.659 & 0.656 & 0.477 & 0.671 & 0.666 & 0.671 & 0.629 \\
\textbf{Qwen3-8B} & \underline{\textbf{0.709}} & \underline{\textbf{0.649}} & 0.590 & \underline{\textbf{0.665}} & \underline{\textbf{0.669}} & 0.542 & \underline{\textbf{0.688}} & \underline{\textbf{0.670}} & \underline{\textbf{0.697}} & \underline{\textbf{0.653}} \\
\textbf{DeepSeek-R1-Distill-Llama-8B} & 0.693 & 0.628 & 0.503 & 0.656 & 0.653 & 0.537 & 0.663 & 0.660 & 0.674 & 0.630 \\
\textbf{Llama-2-7b-chat-hf} & 0.698 & 0.638 & \underline{\textbf{0.658}} & 0.654 & 0.656 & 0.546 & 0.669 & 0.661 & 0.679 & 0.651 \\
\bottomrule
\end{tabular}
\caption{\label{tab2}Main results of each trait. The highest score in each column is bolded and underlined.}
\end{table*}

\section{Results and Discussion}
\subsection{Main Result}
The main results of our method on each prompt and each trait are shown in Table \ref{tab1} and \ref{tab2}. The results show that using linear probes to fit activations of LLMs demonstrates superior performance on a cross-prompt automated essay scoring task. Specifically, using activations from Qwen3-8B and linear probe achieved optimal results, with an average improvement of 9.55\% for prompts and 9.93\% for traits compared to the KAES model. Analyzing the results for each trait, we can see that the QWK scores on the Holistic are not very different for all the models involved in the comparison. The main differences in performance arise from other traits besides Holistic, such as Sentence Fluency and Conventions.

It is worth noting that since we directly search for the optimal head on the test set, the comparison with the baseline does not imply that our method outperforms the baseline. The results presented in Tables \ref{tab1} and \ref{tab2} demonstrate that the activations of LLMs possess strong discriminative power in cross-prompt essay scoring task. This suggests that valuable scoring-relevant information is already encoded in the model’s intermediate representations even without fine-tuning.

A comparison of the results from different models reveals that different large language models have different abilities to evaluate essays. Comparing models with similar model size (DeepSeek-R1-Distill-Llama-8B, Qwen3-8B, and Llama-2-7b-chat-hf), DeepSeek-R1-Distill-Llama-8B performs the worst, with an average decrease of 3.52\% for prompts and 3.52\% for traits compared to Qwen3-8B.

We further explored the effect of model size on the results and found that the larger the model size, the better the results obtained. Comparing the same series of models (Qwen3-8B, Qwen3-4B, Qwen3-1.7B, Qwen3-0.6B) with different model sizes, Qwen3-0.6B performs the worst. As the model size grows, the results on each prompt and each trait get progressively better. Although Qwen3-4B is only half the size of DeepSeek-R1-Distill-Llama-8B, its average QWK score is only 0.01 lower than that of DeepSeek-R1-Distill-Llama-8B. This suggests that choosing a reasonable model is better than choosing a model with a large number of parameters.

Besides, we found that the QWK scores for prompt 7 were low under cross-prompt training, so we tried to use Llama-2-7b-chat-hf to remove the data for prompt 7 during training, and the results are shown in Table \ref{tab3}. After removing the data for prompt 7 during training, the majority of QWK scores were improved. The QWK scores for prompt 8 increased the most, by 23.46\% on Conventions, and the QWK scores for prompt 3 decreased the most, by 2.81\% on Content. Overall, excluding prompt 7 from the training data improves the performance of cross-prompt essay scoring on most of the remaining prompts.

\begin{table}
\setlength{\tabcolsep}{4.5pt}
\begin{tabular}{l|cccc|c}
\toprule
& \textbf{Holistic} & \textbf{Content} & \textbf{Org} & \textbf{Conv} & \textbf{P-AVG} \\
\midrule
\textbf{P1} & -1.39\%  & 0.45\% & 3.11\% & 0.80\% & 0.48\% \\
\textbf{P2} &  2.67\%  & 2.17\% & 3.90\% & 5.80\% & 2.39\% \\
\textbf{P3} & -1.24\%  & -2.81\% & - & - &  -0.82\% \\
\textbf{P4} &  0.83\%  & -1.08\% & - & - & -0.05\% \\
\textbf{P5} &  0.20\%  & -0.30\% & - & - & -0.02\% \\
\textbf{P6} &  2.54\%  &  3.19\% & - & - & 1.16\% \\
\textbf{P8} &  -0.62\% & 13.99\% & 2.58\% & 23.46\% & 5.83\% \\
\midrule
\textbf{T-AVG} & 0.64\% & 1.19\% & 2.66\% & 7.81\% &  \\
\bottomrule
\end{tabular}
\caption{\label{tab3}Relative improvement of removing prompt 7 data from training using Llama-2-7b-chat-hf. P-AVG refers to the average QWK across all traits for each prompt. T-AVG refers to the average QWK across all prompts for each trait.}
\end{table}

\subsection{Performance Under Different Probes}
The results presented above showed that fitting simple linear probes to the activations of LLMs can achieve good performance in cross-prompt essay scoring task. To verify the reliability of the linear probe, we additionally fitted one-layer multilayer perceptions (MLPs) with ReLU non-linearities as nonlinear probes on Llama-2-7b-chat-hf, Qwen3-8B, and DeepSeek-R1-Distill-Llama-8B. The nonlinear probe is formulated as:
\begin{equation}
    \boldsymbol{y_{l,h}} = \hat{\boldsymbol{W}_{2}} \text{ReLU}
    (\hat{\boldsymbol{W}_{1}}\boldsymbol{X_{l,h}} + b_1)+b_2
    \tag{2}
\end{equation}
Where $\hat{\boldsymbol{W}}_{1}$, $\hat{\boldsymbol{W}}_{2}$, $b_1$, and $b_2$ are the weight matrices and bias vectors respectively. Following \cite{gurnee2023language}, we trained the nonlinear probes with a hidden layer size of 256, using the AdamW optimizer with a weight decay of 0.1 and Mean Squared Error (MSE) as the loss function. Given the large volume of training data in the cross-prompt automated essay scoring task, we set the batch size to 2048.

The results are shown in Tables \ref{tab4} and \ref{tab5}. For each LLM evaluated, linear probing consistently outperforms nonlinear probing. Despite achieving similar performance to linear probes on Holistic and Content, nonlinear probes perform markedly worse on the remaining traits, even falling below the results of PAES. The results indicate that, to a certain degree, traits relevant to essays are encoded in a linearly separable manner within large language models.

\begin{table*}[t]
\centering
\setlength{\tabcolsep}{5.2pt} 
\begin{tabular}{lccccccccc}
\toprule
\textbf{Model} & \textbf{P1} & \textbf{P2} & \textbf{P3} & \textbf{P4} & \textbf{P5} & \textbf{P6} & \textbf{P7} & \textbf{P8} & \textbf{AVG} \\
\midrule
Qwen3-8B+\textit{Ridge} & 0.664 & 0.679 & 0.676 & 0.718 & 0.701 & 0.678 & 0.521 & 0.598 & 0.654  \\
DeepSeek-R1-Distill-Llama-8B+\textit{Ridge} & 0.648 & 0.642 & 0.651 & 0.694 & 0.700 & 0.657 & 0.478 & 0.578 & 0.631 \\
Llama-2-7b-chat-hf+\textit{Ridge} & 0.665 & 0.663 & 0.667 & 0.695 & 0.701 & 0.656 & 0.558 & 0.614 & 0.652 \\
\midrule
Qwen3-8B+\textit{MLP} & 0.486 & 0.378 & 0.633 & 0.670 & 0.683 & 0.640 & 0.494 & 0.449 & 0.554 \\
DeepSeek-R1-Distill-Llama-8B+\textit{MLP} & 0.473 & 0.361& 0.572 & 0.557 & 0.671 & 0.576 & 0.397 & 0.394 & 0.500 \\
Llama-2-7b-chat-hf+\textit{MLP} & 0.590 & 0.453 & 0.626 & 0.650 & 0.683 & 0.595 & 0.546 & 0.552 & 0.587 \\
\bottomrule
\end{tabular}
\caption{\label{tab4}Results for different types of probes on each prompt.}

\smallskip
\smallskip
\smallskip

\centering
\setlength{\tabcolsep}{3pt}  
\begin{tabular}{@{}lcccccccccc@{}} 
\toprule
\textbf{Model} & \textbf{Holistic} & \textbf{Content} & \textbf{Org} & \textbf{WC} & \textbf{SF} & \textbf{Conv} & \textbf{PA} & \textbf{Lang} & \textbf{Nar} & \textbf{AVG} \\
\midrule
Qwen3-8B+\textit{Ridge} & 0.709 & 0.649 & 0.590 & 0.665 & 0.669 & 0.542 & 0.688 & 0.670 & 0.697 & 0.653 \\
DeepSeek-R1-Distill-Llama-8B+\textit{Ridge} & 0.693 & 0.628 & 0.503 & 0.656 & 0.653 & 0.537 & 0.663 & 0.660 & 0.674 & 0.630 \\
Llama-2-7b-chat-hf+\textit{Ridge} & 0.698 & 0.638 & 0.658 & 0.654 & 0.656 & 0.546 & 0.669 & 0.661 & 0.679 & 0.651 \\
\midrule
Qwen3-8B+\textit{MLP} & 0.689 & 0.639 & 0.296 & 0.397 & 0.358 & 0.325 & 0.631 & 0.619 & 0.654 & 0.512 \\
DeepSeek-R1-Distill-Llama-8B+\textit{MLP} & 0.668 & 0.549 & 0.295 & 0.386 & 0.362 & 0.260 & 0.550 & 0.508 & 0.603 & 0.464 \\
Llama-2-7b-chat-hf+\textit{MLP} & 0.691 & 0.622 & 0.575 & 0.482 & 0.448 & 0.407 & 0.612 & 0.583 & 0.634 & 0.561 \\
\bottomrule
\end{tabular}
\caption{\label{tab5}Results for different types of probes on each trait}
\end{table*}

\subsection{Visualization}
In this section, we present visualizations to better understand the distribution of activations and the performance of attention heads. We used the t-SNE technique \cite{maaten2008visualizing} to visualize the activation distribution and present the QWK scores as heatmaps. 

\begin{figure*}[t]
\centering
\begin{subfigure}[t]{0.53\columnwidth}
    \centering
    \includegraphics[width=\linewidth]{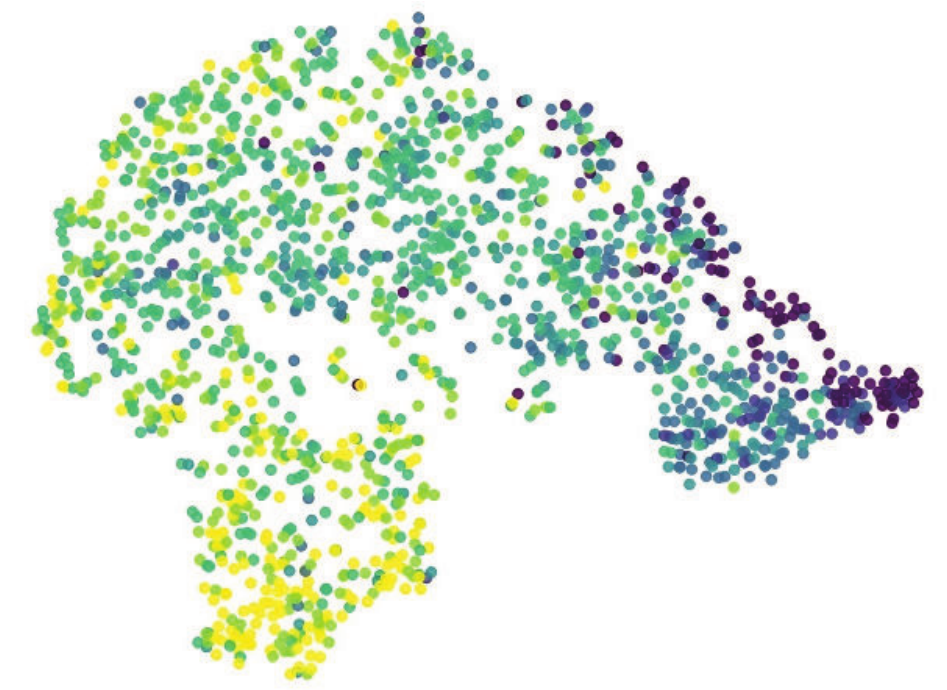}
    \caption{P7 Content L5 H23}
    \label{P7CONT}
\end{subfigure}
\begin{subfigure}[t]{0.53\columnwidth}
    \centering
    \includegraphics[width=\linewidth]{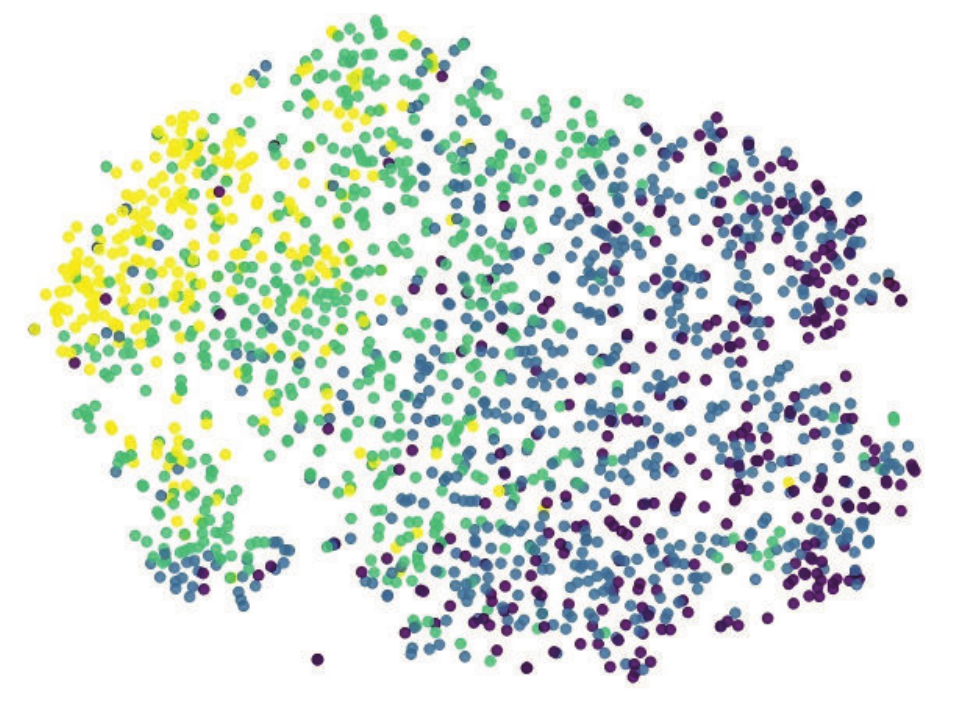}
    \caption{P4 Holistic L11 H16}
    \label{P4HOL}
\end{subfigure}
\begin{subfigure}[t]{0.90\columnwidth}
    \centering
    \includegraphics[width=\linewidth]
    {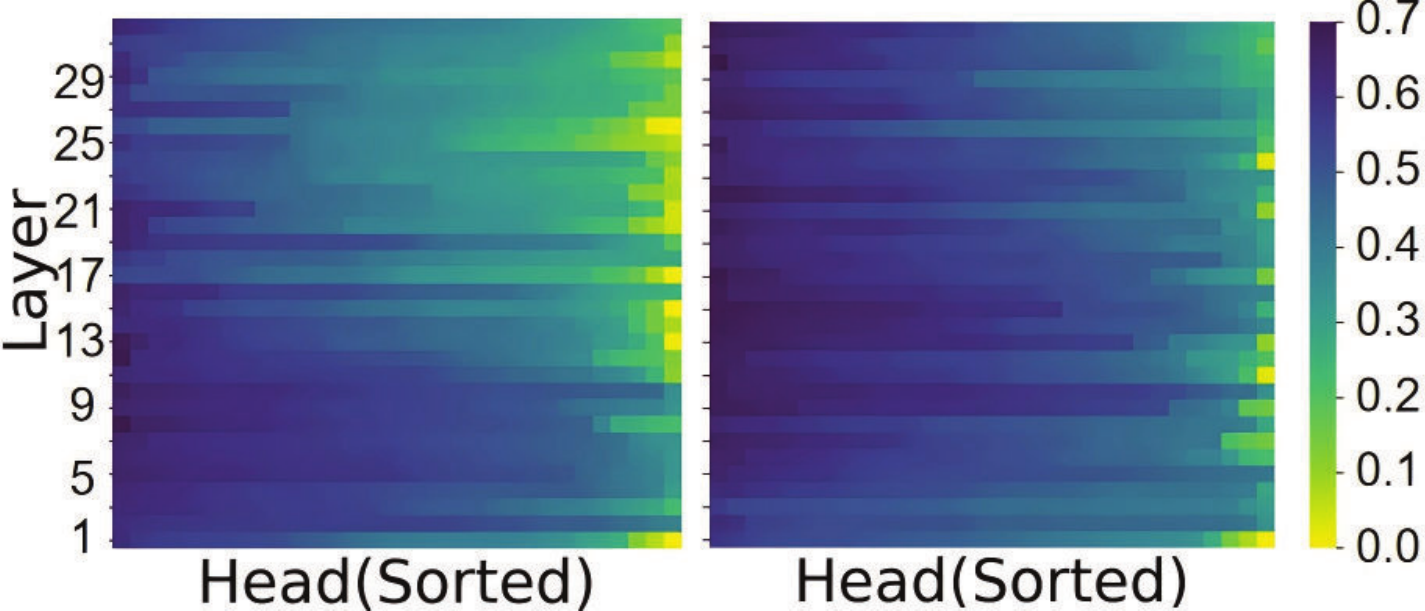}
    \caption{P3 Narrativity(left) and P5 Content(right)}
    \label{P3NAR}
\end{subfigure}
\caption{Visualization of activations and heatmaps of QWK scores. \ref{P7CONT} and \ref{P4HOL} represent the distributions of activations after t-SNE dimensionality reduction. \ref{P3NAR} presents two heatmaps of QWK scores under specific prompts and traits.} 
\end{figure*}

The distribution of activations under the corresponding prompts and traits in the Llama-2-7b-chat-hf is illustrated in Figure \ref{P7CONT} and \ref{P4HOL}. We can see that the 128-dimensional activations show a certain regularity after t-SNE dimensionality reduction, and the activations with high scores and those with low scores are clustered in different regions, which indicates that the activations exhibit strong discriminative power concerning these traits.

Some heatmaps of QWK scores for each prompt and trait are shown in Figure \ref{P3NAR}. Overall, the activations of most attention heads in LLMs can effectively distinguish different traits of the essays. However, for certain traits under specific prompts, the activations in the lower layers (Layers 1–5) and the upper-middle layers (Layers 21–25) exhibit weaker discriminative ability compared to those in the middle layers (Layers 10–16).

\subsection{Sensitivity to Input Content}
When using LLMs to obtain essay activations, a natural question is whether the input content affects the results. To explore this, we designed three experimental settings: removing only the prompt, removing only the instruction, and removing both the prompt and instruction. The same procedure was applied to the Llama-2-7b-chat-hf model using linear probes. The results are presented in Tables \ref{tab6} and \ref{tab7}.

The results show that, when both prompt and instruction were removed and only the essay was retained, the experimental results showed the most substantial performance decay, with the corresponding QWK score decreasing to 0.527 for prompts and traits. The results of removing only the prompt were higher than those from removing only the instruction, and both are lower than the results achieved using complete LLMs' prompts, suggesting that instructions have a greater impact on evaluation essays for large language models compared to prompts. Only when prompt, essay, and instruction are jointly provided as input content to LLMs can LLMs better evaluate the quality of essays.

\begin{table*}[t]
\centering
\setlength{\tabcolsep}{5.2pt} 
\begin{tabular}{lccccccccc}
\toprule
\textbf{Model} & \textbf{P1} & \textbf{P2} & \textbf{P3} & \textbf{P4} & \textbf{P5} & \textbf{P6} & \textbf{P7} & \textbf{P8} & \textbf{AVG} \\
\midrule
Llama-2-7b-chat-hf+\textit{all} & 0.665 & 0.663 & 0.667 & 0.695 & 0.701 & 0.656 & 0.558 & 0.614 & 0.652 \\
Llama-2-7b-chat-hf+\textit{w/o p} & 0.581 & 0.573 & 0.632 & 0.641 & 0.674 & 0.602 & 0.383 & 0.574 & 0.576 \\
Llama-2-7b-chat-hf+\textit{w/o i} & 0.559 & 0.570 & 0.603 & 0.585 & 0.631 & 0.528 & 0.354 & 0.503 & 0.541 \\
Llama-2-7b-chat-hf+\textit{w/o p\&i} & 0.531 & 0.544 & 0.573 & 0.591 & 0.614 & 0.521 & 0.368 & 0.471 & 0.527 \\
\bottomrule
\end{tabular}
\caption{\label{tab6}Results for different input content on each prompt. \textit{all} means input content is complete. \textit{w/o p} means input content without prompt. \textit{w/o i} means input content without instruction. \textit{w/o p\&i} means input content without prompt and instruction.}

\smallskip
\smallskip
\smallskip

\centering
\setlength{\tabcolsep}{3pt}  
\begin{tabular}{@{}lcccccccccc@{}} 
\toprule
\textbf{Model} & \textbf{Holistic} & \textbf{Content} & \textbf{Org} & \textbf{WC} & \textbf{SF} & \textbf{Conv} & \textbf{PA} & \textbf{Lang} & \textbf{Nar} & \textbf{AVG} \\
\midrule
Llama-2-7b-chat-hf+\textit{all} & 0.698 & 0.638 & 0.658 & 0.654 & 0.656 & 0.546 & 0.669 & 0.661 & 0.679 & 0.651 \\
Llama-2-7b-chat-hf+\textit{w/o p} & 0.653 & 0.575 & 0.436 & 0.618 & 0.611 & 0.383 & 0.619 & 0.618 & 0.650 & 0.574\\
Llama-2-7b-chat-hf+\textit{w/o i} & 0.583 & 0.534 & 0.441 & 0.584 & 0.591 & 0.422 & 0.584 & 0.574 & 0.599 & 0.546\\
Llama-2-7b-chat-hf+\textit{w/o p\&i} & 0.579 & 0.525 & 0.412 & 0.569 & 0.576 & 0.374 & 0.560 & 0.552 & 0.597 & 0.527\\
\bottomrule
\end{tabular}
\caption{\label{tab7}Results for different input content on each trait. \textit{all} means input content is complete. \textit{w/o p} means input content without prompt. \textit{w/o i} means input content without instruction. \textit{w/o p\&i} means input content without prompt and instruction.}
\end{table*}

\subsection{Trait-Specific Preferences of Heads}
Through the above experiments, we knew that essays can be evaluated using activations of LLMs, but whether different heads evaluate essays from different perspectives is still unknown. A simple approach is using probes to predict scores for each word, which should be similar if the heads evaluate the essays from the same perspective.

\begin{figure}[tbh]
\centering
\includegraphics[width=\linewidth]{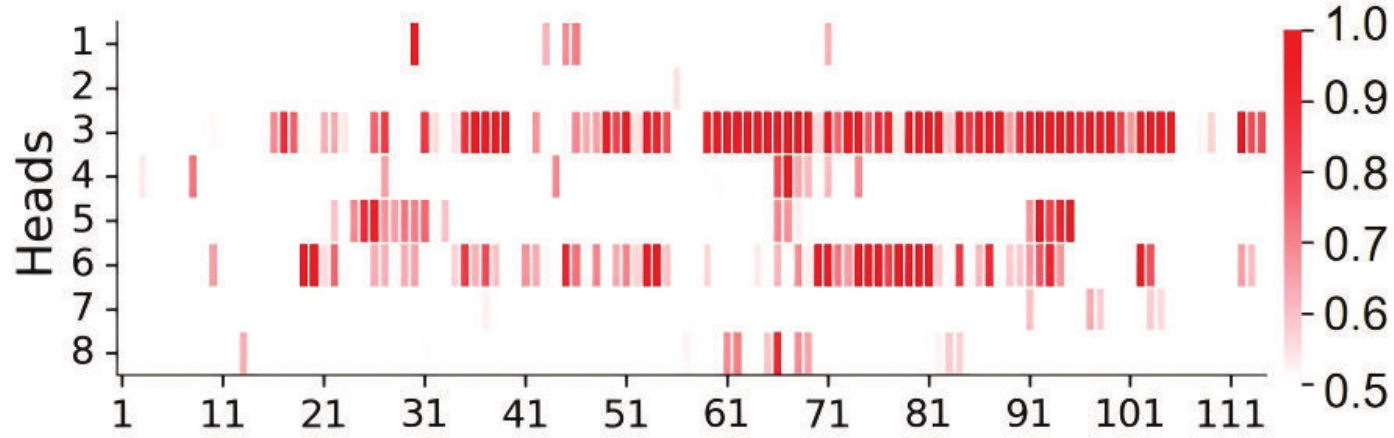}
\caption{The score for each word in the top 8 attention heads. Those greater than 0.5 will be colored.}
\label{head perspective}
\end{figure}

We randomly selected an essay and a trait under a specific prompt, then chose the top 8 attention heads with the highest QWK scores for that prompt-trait pair in the Llama-2-7b-chat-hf model, extracting the activations of all tokens. After that, we used the trained probes to predict scores for each word and visualize the obtained results, which are shown in Figure \ref{head perspective}. We can see that probes from different heads have different prediction scores for the same words, indicating that these heads are evaluating the essay from different perspectives.

\begin{figure*}[tbh]
\centering
\begin{subfigure}[tbh]{\columnwidth}
    \includegraphics[width=\columnwidth]{imgs_pdf/direction/different_prompt/trait_direction_avg_sim_5-5_2_cmyk.pdf}
    \caption{Average cosine similarity between different prompt directions under the same trait. Hol stands for Holistic, Cont stands for Content.}
    \label{trait direction}
\end{subfigure}
\hfill
\begin{subfigure}[tbh]{\columnwidth}
    \includegraphics[width=\columnwidth]{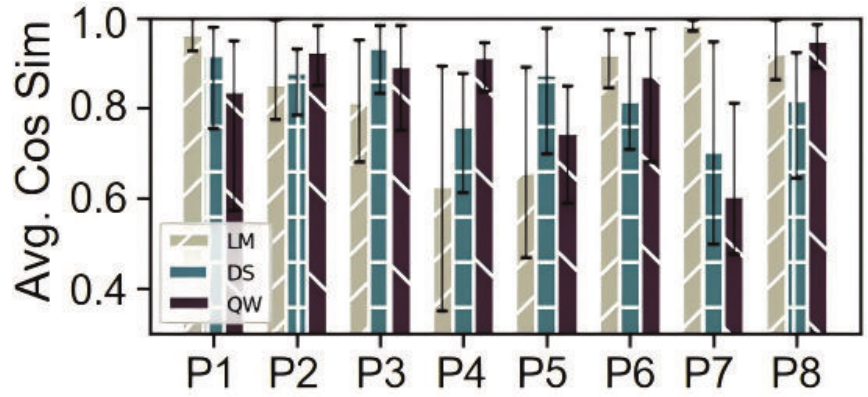}
    \caption{Average cosine similarity between different trait directions under the same prompt.}
    \label{prompt direction}
\end{subfigure}
\caption{Average cosine similarity between different directions. LM stands for Llama-2-7b-chat-hf, DS for DeepSeek-R1-Distill-Llama-8B, and QW for Qwen3-8B.}
\end{figure*}

\section{Direction}
In the section above, we show that the traits of the essay are linearly represented inside the LLMs. In linear representations, directions represent the trend of change of a semantic concept in the embedding space. By analyzing the relationships between different directions, we can identify whether the features in the representation space are independent of each other or are correlated.

\subsection{Direction Calculation}
In the research of representation engineering, most of the studies were binary features or categorical features such as sentiment, topic \cite{turner2023steering}, truth \cite{marks2023geometry, li2023inference}, and safety. Based on the linear hypothesis theory, these studies obtain the corresponding directions by calculating the difference in activations between positive and negative examples:
\begin{align}
\mathbf{v}'_{l,h} &= \frac{1}{N} \sum_{i=1}^{n} \mathbf{a}_{l,h}(P_i^+) - \frac{1}{M} \sum_{j=1}^{m} \mathbf{a}_{l,h}(P_j^-) \tag{3} \\
\mathbf{v}_{l,h} &= \frac{\mathbf{v}'_{l,h}}{\|\mathbf{v}'_{l,h}\|} \tag{4}
\end{align}
where $\mathbf{v}_{l,h}$ is the direction at head $h$ of layer $l$. $N$, $M$ are the number of positive and negative examples, $\mathbf{a}_{l,h}()$ is the activations of the last token at layer $l$ for the given prompt $p$. Unlike binary features, multiple scores for each prompt and trait exist in cross-prompt automated essay scoring task. Therefore, we design the direction of the head $h$ in layer $l$ under the prompt $p$ and trait $t$ as:
\begin{align}
\mathbf{d}_{p,t,i,j} &= \frac{1}{N} \sum_{j=1}^{n} \mathbf{a}_{l,h}(P_{p,t,j}) - \frac{1}{M} \sum_{i=1}^{m} \mathbf{a}_{l,h}(P_{p,t,i}) \tag{5} \\
\mathbf{v}'_{p,t,l,h} &= \sum_{i=score_{p,t}^{Min}}^{score_{p,t}^{Max}-1} \sum_{j=i+1}^{score_{p,t}^{Max}} \mathbf{d}_{p,t,i,j} \tag{6} \\
\mathbf{v}_{p,t,l,h} &= \frac{\mathbf{v}'_{p,t,l,h}}{\|\mathbf{v}'_{p,t,l,h}\|} \tag{7}
\end{align}
where $\mathbf{d}_{p,t,i,j}$ denotes the difference between the activations corresponding to score $i$ and score $j$ under prompt $p$ and trait $t$, $N$ and $M$ denote the number of samples with score $i$ or $j$, prompt $p$, and trait $t$. $P_{p,t,i}$ represents the LLMs' input prompt associated with score $i$, prompt $p$, and trait $t$, $score_{p,t}^{Max}$ refers to the max score and min score under prompt $p$ and trait $t$.

It should be noted that the activations used in this section for our computation of directions were obtained under the full input content, and the QWK scores used are the result of linear probing on the test data.

\subsection{Trait Directions of Different Prompt}
We wanted to know whether the evaluation perspectives of different traits are consistent in a given prompt, so we looked for the heads with the highest average trait QWK scores under each prompt. In these heads, we computed the directions of the different traits and the cosine similarity between them. Results are shown in Figure \ref{trait direction}.

Prompts 1, 2, 3, 6, and 8 exhibit high average directional cosine similarity. Despite significant variations in some models (e.g., prompt 1 in Qwen3-8B and prompt 2 in DeepSeek-R1-Distill-Llama-8B), all three models achieve values above 0.8. The average directional cosine similarity of prompts 4, 5, and 7 varies significantly across different models, with a minimum value of around 0.6 and a maximum value above 0.85. High cosine similarity between feature directions indicates that these traits are not independent in the human evaluation process and often jointly enhance or reduce scores in actual judgments. However, the low cosine similarity between trait directions, such as prompt 7 in Qwen3-8B, suggests that this model can distinguish traits of essays under this prompt more clearly, enabling the generation of separable characteristic representations.

\subsection{Prompt Directions of Different Trait}
We also wanted to know whether the evaluation perspectives of a certain trait are consistent across different prompts within the same head, so we identified the heads with the highest average prompt QWK scores for each trait. In these heads, we computed the directions of the different traits and the cosine similarity between them. Results are shown in Figure \ref{prompt direction}.

For some traits, such as Language and Narrativity, the consistency of cosine similarity values between their prompts was relatively high (e.g., consistently greater than 0.7), suggesting a high degree of consistency in the direction of the cues. In contrast, similarity scores for other traits were very scattered or low, reflecting significant differences in the directions of the corresponding prompts. This phenomenon is consistent with our intuition: some traits are less related to the content of the prompts and are thus more stable across prompts, while other traits, such as content, are more dependent on the specific prompts and have greater directional variability.

By comparing the cosine similarities between prompt directions and trait directions, it can be observed that, in most cases, the similarity between traits is lower than the similarity between prompts. This suggests that, under the same prompt, different traits exhibit a stronger correlation, while different prompts may correspond to more dispersed or differentiated evaluation dimensions. The large range of cosine similarities between directions of different traits under the same prompt, or between different prompts under the same trait, indicates that there are distinct relationships between each pair of traits or prompts.

\section{Conclusion}
In this paper, we explored the discriminative power of activations of large language models in cross-prompt automated essay scoring task. By fitting linear probes to the activations of various LLMs, we achieved excellent results on certain attention heads, verifying the activation's discriminative power in distinguishing essay quality. We further analyzed the factors influencing activation discriminative power and found that choosing a suitable model, rather than merely increasing model size, plays a more crucial role, with instructions also having a significant impact. In addition, by calculating the directions of essays in each trait under different prompts, we found that LLMs can adopt various assessment perspectives according to essay types and traits. In the future, we plan to design more reasonable models with LLM activations as input to improve the accuracy of cross-prompt essay scoring.

\section{Acknowledgments}
This work was supported in part by National Key Program of National Natural Science of China (Grant No.82430108), Guangdong Basic and Applied Basic Research Foundation (Grant No.2024A1515010229), Key Laboratory of Smart Education of Guangdong Higher Education Institutes, Jinan University (2022LSYS003).

\bibliography{aaai2026}


\end{document}